\documentclass[runningheads,a4paper]{llncs}

\usepackage[utf8]{inputenc}

\usepackage{graphicx}
\usepackage{pgfplots}
\pgfplotsset{compat=1.8}
\usetikzlibrary{shapes,arrows}

\usepackage{amsmath}
\usepackage{amsfonts}
\usepackage{amssymb}
\usepackage{array}
\usepackage{mathabx} 
\usepackage{algorithm}
\usepackage{algorithmic}
\usepackage{mathrsfs}

\usepackage{cite}
\usepackage{enumerate}
\usepackage{url}
\usepackage{lipsum}
\usepackage{booktabs} 

\usepackage{dsfont}
\usepackage{latexsym}
\usepackage{hhline}
\usepackage{color}
\usepackage{textcomp}
\usepackage{hyperref}

\makeatletter
\newcommand{\argmin}{\text{argmin}}

\newcommand{\tens}[1]{\boldsymbol{\mathcal{#1}}}

\newcommand{\tT}{\tens{T}}

\definecolor{gris}{gray}{0.90}
\definecolor{gris25}{gray}{0.90}
\definecolor{americanrose}{rgb}{1.0, 0.01, 0.24}
\definecolor{bostonuniversityred}{rgb}{0.8, 0.0, 0.0}
\definecolor{shamrockgreen}{rgb}{0.0, 0.62, 0.38}
\definecolor{selectiveyellow}{rgb}{1.0, 0.73, 0.0}
\definecolor{royalblue}{rgb}{0.25, 0.41, 0.88}
\definecolor{ashgrey}{rgb}{0.7, 0.75, 0.71}
\definecolor{burgundy}{RGB}{159,29,53}
\definecolor{darkgreen}{RGB}{18,53,26}
\definecolor{lightblue}{RGB}{102,217,255}
\definecolor{fakeorange}{RGB}{255,140,102}
\definecolor{arylideyellow}{rgb}{0.91, 0.84, 0.42}
\definecolor{bananayellow}{rgb}{1.0, 0.88, 0.21}
\definecolor{gris_f}{gray}{0.35}
\definecolor{bordure}{rgb}{0.09,0.17,0.68}
\definecolor{aquamarine}{rgb}{0.5, 1.0, 0.83}
\definecolor{apricot}{rgb}{0.98, 0.81, 0.69}
\definecolor{babyblue}{rgb}{0.54, 0.81, 0.94}
\definecolor{uipoppy}{RGB}{225, 64, 5}
\definecolor{uipaleblue}{RGB}{96,123,139}
\definecolor{uiblack}{RGB}{0, 0, 0}
\definecolor{decoda}{RGB}{0,153, 0}
\definecolor{lightgreen}{rgb}{0.56, 0.93, 0.56}
\definecolor{blue_f}{rgb}{0.2, 0.2, 0.6}

\definecolor{lightgreen}{rgb}{0.56, 0.93, 0.56}

\urldef{\mailsa}\path|{jeremy.cohen}@umons.ac.be|    
\newcommand{\keywords}[1]{\par\addvspace\baselineskip
\noindent\keywordname\enspace\ignorespaces#1}

\toctitle{Nonnegative PARAFAC2: a flexible coupling approach}

\begin{document}

\mainmatter  

\title{Nonnegative PARAFAC2: a flexible coupling approach}

\titlerunning{Nonnegative PARAFAC2: a flexible coupling approach}

%

\author{Jeremy E. Cohen%
\thanks{Research funded by F.R.S.-FNRS incentive grant for scientific research n$^\text{o}$ F.4501.16.}%
\and Rasmus Bro}

\institute{Departement of Mathematics and Operational Research, \\ Rue de Houdain 9, Facult\'e polytechnique, Universit\'e de Mons
 \\
Departement of Food Science,\\
Rolighedsvej 26
1958 Frederiksberg C, University of Copenhagen\\
\mailsa}

%
%

\maketitle

\begin{abstract}
Modeling variability in tensor decomposition methods is one of the challenges
of source separation. One possible solution to account for variations from one
data set to another, jointly analysed, is to resort to the PARAFAC2 model.
However, so far imposing constraints on the mode with
variability has not been possible. In the following manuscript, a relaxation
of the PARAFAC2 model is introduced, that allows for imposing nonnegativity
constraints on the varying mode. An algorithm to compute the proposed flexible
PARAFAC2 model is derived, and its performance is studied on both synthetic and
chemometrics data.

    \keywords{PARAFAC2, nonnegativity constraints, flexible coupling}
\end{abstract}

\section{Introduction}

The PARAFAC2 model is an interesting alternative to the more widespread PARAFAC
model~\cite{harshman1972parafac2}. As opposed to PARAFAC, it allows for
non-linearities such that the data need not behave according to a low-rank
trilinear model. In fact, it can even handle sub-matrices (slabs) of varying
length. This is often useful for example when one of the modes is a time
mode~\cite{wise2001application,skov2005new}. One of the prime uses of PARAFAC2
is in the resolution of chromatographic
data~\cite{amigo2008solving,johnsen2017gas,garcia2004building}.

The PARAFAC2 model has shown to have a remarkable ability to resolve
complicated chromatographic data. A typical three-way dataset will have one
mode made up of the various physical samples measured. These could e.g. be
different milk samples.  Another mode will be reflecting elution time which is
a physical separation of the sample over time.  The last mode refers to the
spectral detection such as mass spectrometry --- which represents the actual
measurement of a mass spectrum at each time point for each sample. A successful
PARAFAC2 model will provide loading matrices for each mode that will ideally
represent the concentrations of the compounds measured; the corresponding
elution profiles and the corresponding pure analyte spectra. Usually, the
PARAFAC2 model is only applied to a narrow time interval as for example a
timespan of a few overlapping peaks that are hard to separate without the use
of PARAFAC2.  

In the context of chromatographic data, the ‘selling point’ of PARAFAC2 is that
it allows the elution profile of a given chemical to be different in each
experiment. If chromatographic data would be modelled with a PARAFAC model and
most other conventional curve resolution methods, they would require a given
chemical to have the same elution shape in every sample.  Unfortunately, that
is almost never the case. There will often be retention time shifts and other
shape changes that makes it impossible to model the data with a conventional
approach. The PARAFAC2 model, though, can handle this type of artefacts quite
well.

In many cases, it is desired that the parameters are constrained to be
nonnegative. Most notably because ideally, concentrations, elution profiles and
spectra are nonnegative. Unfortunately, it is not hitherto possible to
constrain all the parameters to be nonnegative. The ‘elution time’ mode of the
PARAFAC2 model is estimated implicitly as a product of two matrices and so far
no algorithms have been presented that allows imposing nonnegativity on the
product of those two matrices. In this paper, we will develop such an
algorithm. In the first section, the PARAFAC2 model is cast as a coupled matrix
factorization model, which is used in section~\ref{sec:FP2NN} to derive an
algorithm for computing Flexible PARAFAC2 with nonnegativity constraints.
Finally, Section~\ref{sec:Exp} shows the performance of the proposed method on
both synthetic and real world data.

\section{Reminders on the PARAFAC2 model}\label{sec:P2}
The PARAFAC2 model was first introduced by Harshman in the context of
phonetics~\cite{harshman1972parafac2}. In his work, Harshman looked for a way to
factorize simultaneously several matrices given that one factor was almost the
same, but not exactly, in all those matrices. He thus imposed a linear
transformation as a coupling relationship between the similar factors. However
using a generic linear coupling model adds too many parameters, and to ensure
identifiability of both the factors and the coupling matrices, orthogonality
constraints were imposed. This leads to the following PARAFAC2 model:
\begin{equation}
    \begin{array}{l}
        M_{k} = AD_{k}B_{k}^{T} + E_{k} \\
        B_{k} = P_{k}B^{\ast} \\
        P_{k}^{T}P_{k} = I_{r} 
    \end{array}\label{eq:p2},
\end{equation}
where $B^\ast$ is a $r\times r$ matrix of coefficients, $D_{k}$ is a $r\times r$ diagonal
matrix, $P_{k}$ is a $m_{k}\times r$ left-orthogonal matrix and $E_{k}$ is a
$n\times m_{k}$ residual error matrix. Here the
coupled matrices are the $B_k$, and the coupling matrices, the $P_k$.

Another way to understand PARAFAC2, more widely used in the tensor community,
is to cast it as a relaxation of the PARAFAC model. Indeed, stacking matrices $M_k$
into one large tensor $\tT$, the PARAFAC2 model yields:
\begin{equation}
    \begin{array}{l}
        T_{ijk} = \sum\limits_{p=1}^{r}{A_{ip}B^{(k)}_{jp}C_{kp}} \\
        \sum\limits_{p=1}^{r}{ B^{(k)}_{j_{1}p} B^{(k)}_{j_{2}p}} =
        \sum\limits_{p=1}^{r}{B^{\ast}_{j_{1}p}B^{\ast}_{j_{2}p}} \quad \forall j_{1},j_{2},k
    \end{array},
\end{equation}
where $C$ is obtained by stacking the diagonals of $D_k$ in rows. One can
observe that contrary to the PARAFAC model, the $B$ factor is allowed to vary
for each slice $k$. This variation is controlled by the inner products stored
in $B^\ast$ and kept constant through $k$. As a matter of fact, the
orthogonality constraints on the $P_k$ matrices are equivalent to imposing a
shared Gramian matrix for all $B_k$, that is $B_k^T B_k = {B^{\ast}}^T B^\ast$ for
all $k$. The power of PARAFAC2 comes from the fact that this constraint is
implicit, and may give birth to a wide range of variability among the $B_k$
while maintaining an overall coupling structure. In contrast, other similar
models like Shift-PARAFAC impose a coupling constraint
in an explicit fashion that may be too specific and difficult to
implement~\cite{harshman2003shifted,morup2008shift}.

To identify the parameters of the (unconstrained) PARAFAC2 model, the following optimization
problem needs to be solved:
\begin{equation}\label{eq:optp2}
    \begin{array}{l}
    \underset{A,D_{k},P_{k},B^\ast}{\argmin} \sum\limits_{k=1}^{K}{\| M_k -
    AD_{k}\left(P_{k}B^{\ast}\right)^T \|_F^2} \\
    \text{so that } P_{k}^{T}P_{k} = I
\end{array}.
\end{equation}
An efficient alternating algorithm to solve~\eqref{eq:optp2} has been introduced
in~\cite{kiers1999parafac2}. It relies on the fact that if the $P_k$ matrices are known,
then multiplying each data slice $M_k$ by $P_k$ on the right, the PARAFAC2 model becomes a
PARAFAC model with second mode factor $B^\ast$. Therefore, an alternating
algorithm may first estimate $P_k$ fixing the other parameters, then
pre-process the data by multiplying each slice with $P_k^T$, and then use a few
step of an algorithm to compute PARAFAC, for instance Alternating Least
Squares~\cite{comon2009tensor}. The estimation of the orthogonal coupling matrices is
easily obtained with SVD, knowing that the solution of
\begin{equation}
    \begin{array}{l}
        \underset{P\in\mathds{R}^{m\times r}}{\argmin} \| M - PX \|^2_F \\
    \text{such that } P^{T}P = I
    \end{array}
\end{equation}
is given by $P = U(:,1:r){V(:,1:r)}^T$, where $[U,S,V]$ is the Singular Value
Decomposition of $MX^T$.

\section{About exact nonnegative PARAFAC2}

Imposing nonnegativity on the $B$ mode in the PARAFAC2 model is known to be a
difficult problem and no solver actually implements it currently. Let us show
rapidly why it is not straightforward, but still feasible, to impose
nonnegativity within the algorithmic framework described above, that is when
estimating $P_k, A, B^\ast$ and $C$ alternatively.

Clearly, imposing nonnegativity on $B^\ast$ --- which would be possible since
nonnegativity is well understood for PARAFAC --- does not guaranty that the
reconstructed $B_{k} = P_{k}B^\ast$ are themselves nonnegative. Therefore, the
following set of constraints has to be imposed on $P_k$ and $B^\ast$ in the
PARAFAC2 model: 
\begin{equation}\label{eq:constr}
    P_{k}B^\ast \geq 0 \quad \forall k \in [1,l],
\end{equation}
which requires to modify the estimation procedures of both $P_k$ and $B^\ast$.

\subsection{Estimating the orthogonal coupling matrices}
The estimation of the orthogonal matrices $P_k$ is a crucial step in the ALS
algorithm which can be done slice by slice. The following optimization problem
is solved:
\begin{equation}\label{eq:PkNN}
    \begin{array}{l}
        \underset{P_k \in \mathds{R}^{m_k\times r}}{\argmin}    \| M_k - AD_{k}{\left(P_{k}B^{\ast}\right)}^T
    \|_F^2 \\
    \text{so that } P_{k}^{T}P_{k} = I, \quad P_{k}B^\ast \geq 0
    \end{array}.
\end{equation}

Without nonnegativity constraints, $P_k$ is computed using the Singular
Value Decomposition (SVD). Sadly such a simple procedure cannot be used
anymore in order to build a converging optimization algorithm because of the
nonnegativity constraints. This optimization problem is reminiscent of the
Orthogonal Nonnegative Matrix Factorization problem~\cite{pompili2014two} which is
difficult to solve. 

\subsection{Estimating the latent factor} 
Supposing matrices $P_k$ have been
computed in a previous step, after the data matrices $M_k$ have been processed
by multiplying them with $P_{k}^T$, the second mode variable in the PARAFAC
model becomes $B^\ast$. 

Within the framework of alternating optimization that we want to use
here\footnote{Alternating optimization may be avoided using an all-at-once
method but the problem of satisfying the nonnegativity
constraints still remains.}, knowing the current estimates for $A$ and $C$, the
following optimization problem is to be solved:

\begin{equation}
    \begin{array}{l}
        \underset{B^\ast\in\mathds{R}^{r\times r}}{\argmin} ~\frac{1}{2}\| T_{\left[2\right]} - B^\ast {\left( A\odot C
\right)}^T 
    \|_F^2 \\
    \text{s.t. } P_{k}B^\ast \geq 0 \quad \forall k \in [1,l]
\end{array}\label{eq:optpb}.
\end{equation}

A possible approach to our problem would be to solve the exact nonnegative
least squares using the Kronecker structure of the problem. This is by no means an easy
task, and we could find no other work related to this issue.
Another approach would be to use a projected gradient,
but a projector on the
constraint space would then be needed, which is not known in closed form.

As a consequence, since both the estimation of $P_k$ and $B^\ast$ are
cumbersome, the algorithm implementing the methods described above proved to be
quite slow and very sensitive to initialization, making it mostly useless in
practice. That is the reason why the flexibly coupled PARAFAC2 is introduced in
the next section.

\section{A flexible PARAFAC2 model}\label{sec:FP2NN}

As described in Section~\ref{sec:P2}, the PARAFAC2 model can be understood as a
coupled matrix low rank factorization, where the coupled factors $B_k$ are
constrained to have the same inner products. The difficulty of working with
constrained PARAFAC2 is that, by parameterizing each $B_k$ as $P_{k}B^\ast$,
constraints on the coupled mode are imposed on a product of two blocks of
variables. In particular the $P_k$ matrices are already constrained to be
orthogonal.

Moreover, even though PARAFAC2 is less constrained than PARAFAC and has
therefore been applied in context of subject variability, it makes the
important underlying assumption that all the columns
of $B_k$ are transformed similarly, by opposition to component by component
transformation found in other related models like Shift-PARAFAC. For instance,
in the context of Gas
Chromatography---Mass Spectroscopy, from one batch to another, elution profiles
change in a slightly unpredictible manner, and their inner products are not exactly
constant over the batches.  Relaxing the hard coupling constraint in PARAFAC2
could allow for a better fitting of the PARAFAC2 in difficult cases.

For both those reasons, it makes sense to introduce a Flexible PARAFAC2 model,
where the coupled factors $B_k$ are no longer parameterized, but instead
constrained to be close to $P_{k}B^\ast$. Formally, the Flexible
PARAFAC2 model can be cast as follows: 
\begin{equation}
    \begin{array}{l}
        M_{k} = AD_{k}B_{k}^{T} + N_{k} \\
        B_{k} = P_{k}B^{\ast} + \Gamma_{k} \\
        P_{k}^{T}P_{k} = I_{r} \\
        \|A(:,i)\|_2 = 1 \quad \forall i\in\{1..r\}
        \|B^{\ast}(:,i)\|_2 = 1 \quad \forall i\in\{1..r\}
    \end{array}\label{eq:fp2},
\end{equation}
where $\Gamma_{k}$ is an coupling error matrix.
This kind of flexible coupling have been introduced
in~\cite{farias2015exploring} and under Gaussianity assumption for both model and coupling
errors, a Maximum A Priori estimator of the different variables can be easily obtained
by solving an optimization problem, here cast with nonnegativity constraints:
\begin{equation}\label{eq:optfp2}
    \begin{array}{l}
        \underset{A,B_{k},B^{\ast},P_{k},D_{k}}{\argmin} \sum\limits_{k=1}^{K}
        \|M_{k} - AD_{k}B_{k}^{T} \|_F^2 + \mu_k \|B_{k} - P_{k}B^{\ast} \|_F^2 \\
        \text{so that } A\geq0, B_k \geq 0, D_k \geq 0, \|A(:,i)\|_2 = 1,
        \|B^{\ast}(:,i)\|_2 = 1  \quad
        \forall i\in\{1,..,r\}
    \end{array},
\end{equation}
where $\mu_k$ is a collection of regularization parameters controlling the
distance between the factors $B_k$ and their coupled counterparts
$P_{k}B^\ast$. If noise levels on each data slice $M_k$ are available, they can
be added as a normalization constant in front of the data fitting terms. Note
that the normalization of $A$ and $B^{\ast}$ in equation~\eqref{eq:fp2} is important, otherwise
the regularization parameters $\mu_k$ and the latent factor $B^\ast$ are
defined up to scaling and that makes the coupling terms difficult to interpret.

The main advantage of solving~\eqref{eq:optfp2} over~\eqref{eq:optp2} is that
the nonnegativity constraints now apply directly on factors $B_k$. In an
alternating optimization scheme, alternating over variables $A,D_k,B_k,P_k$ and
$B^\ast$, the coupled factors can be estimated with a simple nonnegative least squares algorithm, for
instance~\cite{gillis2012accelerated}. The estimates for $P_k$ can be obtained using
SVD, and computing $B^\ast$ is a least squares problem. Therefore deriving an
alternating optimization algorithm as the suggested Algorithm~\ref{alg:fp2}
is straightforward. Moreover, because each sub-problem in
Algorithm~\ref{alg:fp2} is optimally solved, given that the parameters $\mu_k$
are kept constant, the cost function is guarantied to decrease after each
iteration. Therefore, the proposed algorithm for computing Flexible PARAFAC2 is
guarantied to converge. 

At this stage, the Flexible PARAFAC2 model can be thought of as a relaxation of
the PARAFAC2 model, but it is also possible to interpret~\eqref{eq:optfp2} as a
relaxed optimization problem to solve the exact PARAFAC2 model. Then by
increasing the values of $\mu_k$ during the optimization algorithm,
asymptotically, minimizing~\eqref{eq:optfp2} yields an exactly coupled PARAFAC2
model. As a consequence, introducing flexibility may be understood as an optimization trick
that makes constrained PARAFAC2 easier to compute. Practically, the residual relative coupling errors
$\frac{\|B_k - P_{k}B^\ast \|_F^2}{\| B_k \|_F^2}$ can be monitored so that
when a low value of such error is reached, the regularization parameter $\mu_k$ may stop
increasing to ensure final convergence.\vspace*{-0.5cm}
\begin{algorithm}\label{alg:fp2}
    \begin{algorithmic}
        \STATE{\textbf{INPUT:} Data slices $M_k$, initial guesses for factors
        $A,D_{k},B_{k},P_{k},B^{\ast}$.}
        \STATE{1. Set small initial values for $\mu_k$
        using~\eqref{eq:muinit} and normalize $M_k$ with the total $\ell_2$
    norm of all slices.}
        \WHILE{Stopping criterion is not met}
            \STATE{2. For all $k$, increase $\mu_k$ if necessary}
            \STATE{3. For all $k$, $P_k$ estimation: $ P_k = U(:,1:r){V(:,1:r)}^T$ 
                \\ where  $[U,S,V^T] = \text{SVD}(B_{k}{B^{\ast}}^T)
            $}
            \STATE{4. $B^\ast$ estimation: $
                B^\ast =
                \frac{1}{\sum\limits_{k=1}^{K}{\mu_k}}\sum\limits_{k=1}^{K}{\mu_{k}P_{k}^{T}B_k
            }$ normalized columnwise.}
            \STATE{5. $A$ estimation: $A = \underset{A\geq 0}{\argmin}
                    \sum\limits_{k=1}^{K}{\| M_k - AD_{k}B_{k}^T \|^2_F}
                 $ solved by nonnegative least squares, then normalized
             columnwise.}
            \STATE{6. For all $k$, $B_k$ estimation:  $  B_k = \underset{B_k\geq 0}{\argmin}
                    \| M_k - AD_{k}B_{k}^T \|^2_F + \mu_k \| B_k - P_{k}B^{\ast}
                    \|^2_F
                 $ solved by nonnegative least squares.}
            \STATE{7. For all $k$, $D_k$ estimation:  $D_k = \underset{D_k\geq 0}{\argmin}
                    \| M_k - AD_{k}B_{k}^T \|^2_F
                 $ solved by nonnegative least squares after vectorization.}
            \STATE{8. If this is the first iteration, for all $k$, choose $\mu_k$ so that regularization
            is a certain percent of cost function using~\eqref{eq:muinit}.}
        \ENDWHILE\STATE{\textbf{OUTPUT:} Estimated nonnegative factors $A,D_{k},B_{k}$
        and coupling factors $P_{k},B^{\ast}$.}
    \end{algorithmic}
    \caption{Alternating nonnegative least squares algorithm for solving
    Flexible PARAFAC2 with nonnegativity constraints.}
\end{algorithm}\vspace*{-1.0cm}
\paragraph{Remark}
If the parameters $\mu_k$ increase too fast at the beginning of the algorithm,
then the updates of $B_k$ are mostly driven by the regularization term. In that
case, we observed that the values of $B_k$ do not change much, and the
algorithm ends up in a local minimum where only $A$ and $C$ are optimized.
Therefore, it is important to not increase parameters $\mu_k$ over some
reasonable threshold that depends on the data fitting terms.
For the initial value of $\mu_k$ and their values after the first iteration,
we used respectively 
\begin{equation}\label{eq:muinit}
    \mu_k^{0} = 10^{-1} \frac{\|M_k - A^{0}D^{0}_{k}{B^{0}_{k}}^{T}
    \|_F^2}{\|B^{0}_k \|_F^2} \text{  and  }
    \mu_k^{1} = 10^{-SNR/10} \frac{\|M_k - A^{1}D^{1}_{k}{B^{1}_{k}}^{T} \|_F^2}{\|B^{1}_k -
    P^{1}_{k}{B^{1}}^{\ast} \|_F^2}
\end{equation}
where $A^{0}$ is the initial value of $A$, $A^{1}$ is the estimate of $A$ after the first
iteration and SNR refers to the expected Signal to Noise ratio of the whole
tensor data. These values can be tuned by the user if
necessary. The increase of $\mu_k$ at each iteration is implemented as
a multiplication of the current value by $1.02$ if $\mu_k\leq 10$.

\paragraph{Initialization}
In the experiments conducted in the next section, we used random
factors for initialization. Another possible choice for the factors initial values is
to use the output of a PARAFAC model or to compute  independent nonnegative
matrix factorizations for each slice. In our experiments, all these methods provided good
initialization to the flexible PARAFAC2 model, yet this claim will be rigorously
studied in later research. A good
choice of initial $P_k$ in any case is the zero-padded identity matrix. 

\section{Experiments on synthetic data}\label{sec:Exp}

In this section, we provide experimental proof that the proposed Flexible
PARAFAC2 model allows for imposing nonnegativity constraints on the $B$ mode
while showing performance at least similar to the state-of-the-art PARAFAC2 algorithm
introduced in~\cite{kiers1999parafac2}. Also, we show that the proposed algorithm exhibits
better robustness to random initialization, which in practice means a reduced
number of initial trials is required.

The synthetic data are generated as follows. The dimensions are set to
$[20\times 30\times
20]$ and the rank to $R=3$. The entries of factors $A$ are Gaussian with unit
variance, then clipped to zero to have a sparse factor matrix. The entries of factor
$C$ are drawn from a uniform distribution on $[0,1]$. Both $A$ and $C$ are then
normalized column-wise using the $\ell_2$ norm. In the experiments above, an
i.i.d. Gaussian noise of variance $\sigma^2$ is added to each entry of the
obtained tensor, where $\sigma$ is a parameter of the experiments.

Generating nonnegative factors $B_k$ that have the same Gramiam matrix is not
straightforward. In this manuscript, we used a particular coupling between the
$B_k$ for which the inner products are trivially kept constant over the third mode.
Namely, a first factor $B_1$ is drawn entry-wise using a Gaussian unitary
distribution, clipped to 0 and normalized column-wise, then factors $B_k$ are
obtained by circularly shifting $B_1$ along the grid of indexes. The obtained
model is then actually a Shifted PARAFAC model, which is a particular case of
PARAFAC2 that can be easily generated for simulation purpose. 

The maximum number of iterations is set to 1000, and a stopping criterion based
on the relative error decrease is used.

The experiments are conducted to check the performance of the nonnegative
flexible PARAFAC2 proposed algorithm with respect to the state-of-the-art
algorithm~\cite{kiers1999parafac2}, which does not implement nonnegativity on the coupled mode.
To this end, the following relative error on factors $B_k$ is computed for $N=50$
simulated tensor data drawn with various noise values $\sigma$ ranging from
$5\times 10^{-3}$ to $10^{-5}$:
\begin{equation}
    \frac{1}{K}\sum\limits_{k=1}^{K}{\frac{\|B_k - \left[\widehat{B}_k\right]^+
    \|_F^2}{\|B_k\|_F^2}},
\end{equation}
where all $B_k$ and $\widehat{B}_k$ have been normalized column-wise.

\begin{figure}
        \includegraphics[width=0.48\textwidth]{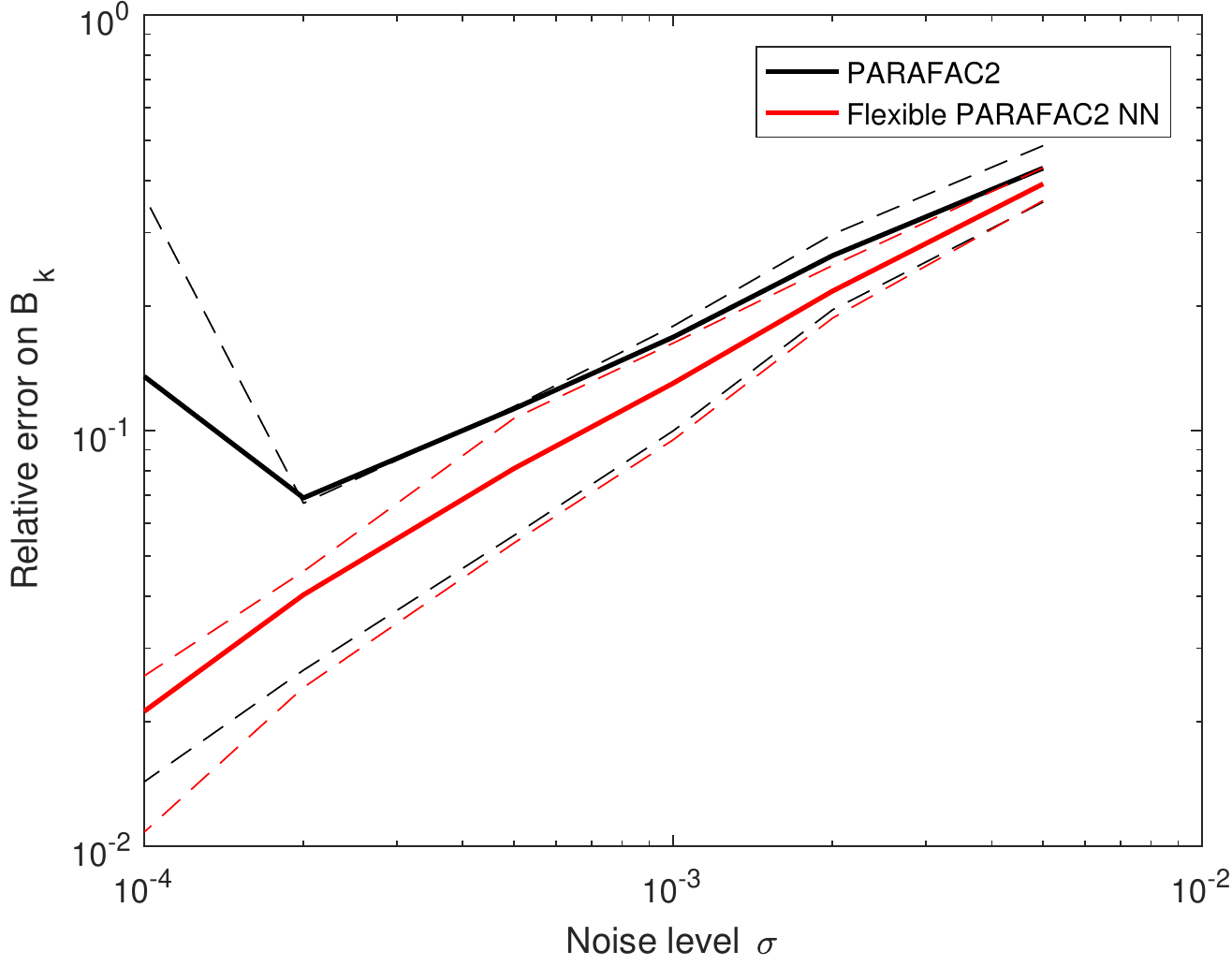}
        \includegraphics[width=0.48\textwidth]{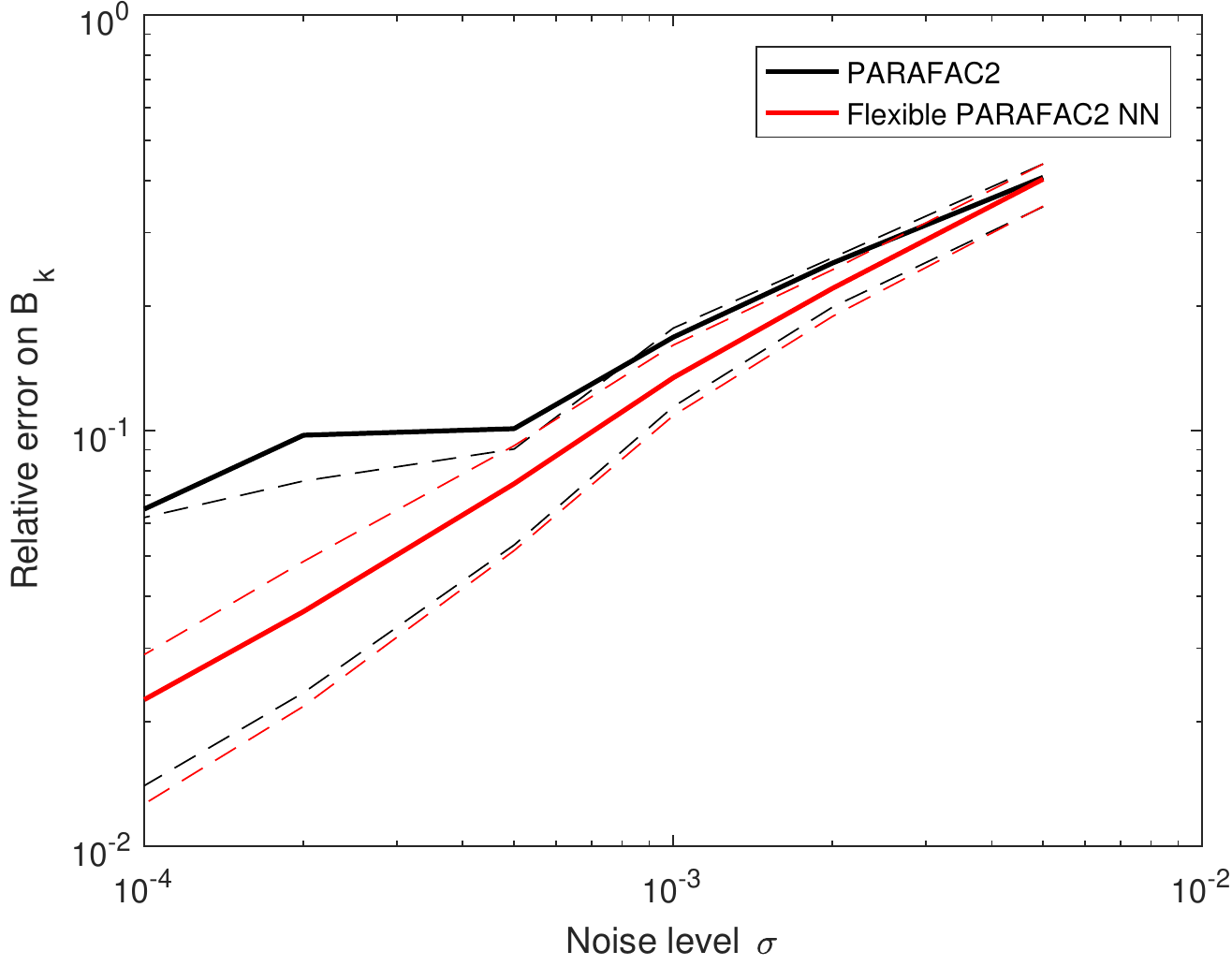}
    \caption{Relative error on $B_k$ for one initialization (left) and
   best of five initializations (right), showing in dotted lines the 20\% and 80\% quantiles.}\label{fig:allinit}
\end{figure}

To study robustness to initialization, Figure~\ref{fig:allinit} exhibits
the error on $B_k$ with both one random initialization and
the best out of five initializations. 

From the results shown in Figure~\ref{fig:allinit}, it
can be concluded that, for the specific choice of shifted coupled factor $B_k$,
the flexible PARAFAC2 best performance is similarly to the state-of-the-art
PARAFAC2 algorithm best performance, with slightly lower estimation error due to the
nonnegativity constraints applied on the $B$ factor. Also, the average
performances are significantly better for the flexible PARAFAC2, and the worst
results are also much closer to the best ones. Therefore, it seems that the
flexible algorithm is more robust to random initializations. 

\section{Experiments on chromatography data}

To further asses the performance of the proposed flexible PARAFAC2 model with
nonnegativity constraints, a Gas Chromatography Mass Spectroscopy interval
is deconvolved, for which the usual PARAFAC2 model produces poor results. The
data come from an analysis of various types of red wine of the type Cabernet
Sauvignon. The analysis was done using headspace GC–MS analysis on a Hewlett
Packard 6890 GC coupled with an Agilent (Santa Clara, California, United
States) 5973 Mass Selective Detector.  More details can be found in the
publication by Ballabio et al~\cite{ballabio2008classification}.

The chosen interval is difficult to decompose since there is supposedly a
double peak in the time elution factors, meaning that the columns of the $B_k$
factor are highly colinear. The rank is expected to be either 3 or 4, so that
both values were used in the comparisons below. Initial factors for the
PARAFAC2 decompositions were drawn from uniform distributions on [0,1]. We
picked the best results out of ten runs for both the unconstrained and flexible
PARAFAC2 algorithms, based on the reconstruction error. 

\begin{figure}
    \includegraphics[width=0.48\textwidth]{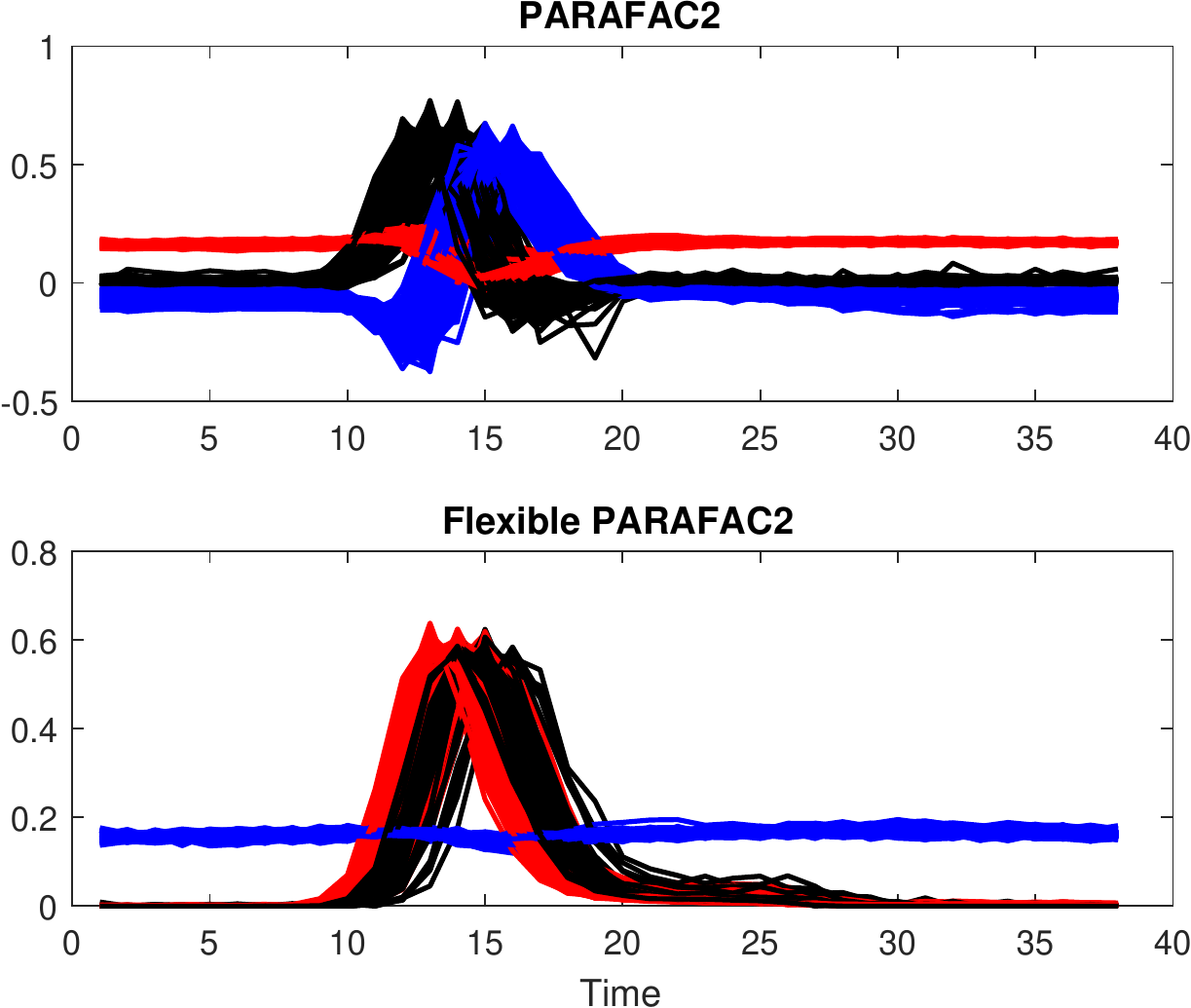}
    \includegraphics[width=0.48\textwidth]{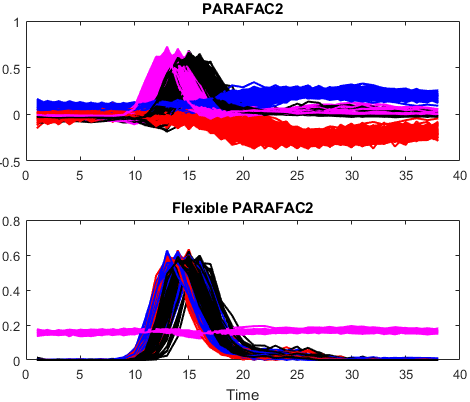}
    \caption{Elution profiles obtained by (top) PARAFAC2 and (bottom) Flexible
    PARAFAC2 with nonnegativity constraints. The rank is set to (left) 3 and
(right) four.}\label{fig:lcms_res}
\end{figure}

Results are presented in Figure~\ref{fig:lcms_res}. First it can be observed
that the elution profiles obtained using PARAFAC2 and flexible PARAFAC2 with
nonnegativity constraints are different. Only the flexible PARAFAC2 outputs
make sense in terms of elution profiles, for both three and four components
models. Moreover, the flexible PARAFAC2 model identifies a very faint third
peak in the elution profiles for a four components model, which is not properly
detected by the PARAFAC2 model without nonnegativity constraints. This means
that the nonnegative flexible PARAFAC2 model has the ability to produce
solutions to some problems the PARAFAC2 model could not solve. It is important
to note, however, that for most other intervals that we studied, both
algorithms performed similarly.

\vspace*{-0.8cm}

\section{Conclusion}
The difficult problem of imposing nonnegative constraints on the coupled mode
in the PARAFAC2 model is tackled in this manuscript. Using a flexible coupling
formalism, the coupled variables and their latent representation are split,
which leads to a simple constrained alternating least squares algorithm that is
easily shown to converge for fixed regularization parameters. Through the
decomposition of both
simulated and gas chromatography mass spectroscopy data, it is shown that
the proposed flexible PARAFAC2 model behaves at worse similarly to the state-of-the-art
PARAFAC2 model, but implementing nonnegativity constraints on all modes and
featuring more robustness to random initialization.
Further works will focus on a more precise analysis of the flexible PARAFAC2
model for solving various problems, and an extension for imposing any
off-the-shelf constraints on the coupled mode.

\section{Acknowledgements}

The authors wish to thank Nicolas Gillis for helpful discussions on
alternatives to the flexible coupling approach for computing nonnegative
PARAFAC2.

\bibliographystyle{plain}

\end{document}